\documentclass[letterpaper]{article} 
\usepackage{aaai2026}  
\usepackage{times}  
\usepackage{helvet}  
\usepackage{courier}  
\usepackage[hyphens]{url}  
\usepackage{graphicx} 
\usepackage{natbib}  
\usepackage{caption} 
\usepackage{algorithm}
\usepackage{algorithmic}

\usepackage{newfloat}
\usepackage{listings}
\DeclareCaptionStyle{ruled}{labelfont=normalfont,labelsep=colon,strut=off} 
\floatstyle{ruled}
\newfloat{listing}{tb}{lst}{}
\floatname{listing}{Listing}
\usepackage{array}
\usepackage{booktabs}
\usepackage{makecell}
\usepackage{amssymb,amsmath}
\usepackage{relsize}
\usepackage{subfigure}

\title{DTGen: Generative Diffusion-Based Few-Shot Data Augmentation for Fine-Grained Dirty Tableware Recognition}
\author {
    Lifei Hao\textsuperscript{\rm 1,\rm 2},
    Yue Cheng\textsuperscript{\rm 1,\rm 2},
    Baoqi Huang\textsuperscript{\rm 1,\rm 2, \rm *},
    Bing Jia\textsuperscript{\rm 1,\rm 2},
    Xuandong Zhao\textsuperscript{\rm 1,\rm 2}
}
\affiliations {
    \textsuperscript{\rm 1}Engineering Research Center of Ecological Big Data, Ministry of Education\\
    \textsuperscript{\rm 2}College of Computer Science, Inner Mongolia University\\
    \textsuperscript{\rm *}Corresponding Author, e-mail: cshbq@imu.edu.cn
}

\usepackage{bibentry}

\begin{document}

\maketitle

\begin{abstract}
Intelligent tableware cleaning is a critical application in food safety and smart homes, but existing methods are limited by coarse-grained classification and scarcity of few-shot data, making it difficult to meet industrialization requirements. We propose DTGen, a few-shot data augmentation scheme based on generative diffusion models, specifically designed for fine-grained dirty tableware recognition. DTGen achieves efficient domain specialization through LoRA, generates diverse dirty images via structured prompts, and ensures data quality through CLIP-based cross-modal filtering. Under extremely limited real few-shot conditions, DTGen can synthesize virtually unlimited high-quality samples, significantly improving classifier performance and supporting fine-grained dirty tableware recognition. We further elaborate on lightweight deployment strategies, promising to transfer DTGen's benefits to embedded dishwashers and integrate with cleaning programs to intelligently regulate energy consumption and detergent usage. Research results demonstrate that DTGen not only validates the value of generative AI in few-shot industrial vision but also provides a feasible deployment path for automated tableware cleaning and food safety monitoring.
\end{abstract}

\begin{links}
    \link{Code\&Datasets}{https://github.com/cyicz123/DTGen.git}
\end{links}

\section{Introduction}
Artificial intelligence (AI) has become a transformative force in food safety monitoring, reshaping traditional paradigms and enabling more reliable hygiene control. Industry reports show that AI applications can reduce contamination risks in processing facilities by up to $30\%$, with the sector experiencing sustained growth \cite{Emergen2024}. By leveraging real-time monitoring, predictive analytics, and automated detection, AI systems mitigate human errors and ensure hygiene compliance across the food production chain \cite{AI_Food_Safety_2023}. Within this context, automated tableware cleaning technology has emerged as a key application scenario, extending from industrial kitchens to smart homes as an essential AI-powered hygiene safeguard.

Despite increasing adoption, current research on tableware hygiene classification remains limited. Commercial dishwashers have begun incorporating AI and computer vision to optimize cleaning \cite{homeconnect_dishwasher_2025,hobart_ai_dishwashing_2025}, yet most academic work is restricted to coarse-grained binary classification (clean vs. dirty) \cite{zhu2021deep}, insufficient for practical intelligent cleaning systems that require recognition of contamination type, severity, and spatial distribution. Compounding this challenge, tableware contamination recognition faces acute data scarcity \cite{platesv2}: the combinatorial diversity of food residues, materials, and contamination patterns makes comprehensive annotation economically prohibitive, resulting in few-shot learning constraints. These issues are mutually reinforcing: coarse-grained recognition limits utility, while data scarcity hinders the development of fine-grained models. Addressing fine-grained tableware contamination recognition under extreme few-shot conditions is thus both theoretically significant and practically urgent.

To this end, we propose DTGen, an end-to-end few-shot data augmentation framework based on generative diffusion models, specifically tailored for fine-grained contamination recognition. DTGen reframes the task as multi-level classification and integrates three key innovations: (i) domain-specialized diffusion model fine-tuning via LoRA, (ii) structured contamination-aware prompt generation, and (iii) cross-modal filtering with CLIP to ensure semantic reliability. With only $40$ annotated images, DTGen generates over $3000$ high-quality synthetic samples. Classifiers trained on this synthetic dataset achieve $93\%$ binary accuracy (a $28\%$ improvement over few-shot baselines and $9\%$ over traditional augmentation with larger datasets) and $86\%$ accuracy on fine-grained three-class tasks, substantially surpassing real-data baselines. The framework thus bridges the critical gap in fine-grained tableware contamination recognition and offers a broadly applicable solution for few-shot anomaly detection in industrial vision inspection.

Our contributions are threefold: (1) We redefine tableware contamination recognition from binary to fine-grained multi-level classification, establishing a foundation for comprehensive automated hygiene monitoring. (2)We adapt state-of-the-art generative AI techniques (LoRA fine-tuning, structured prompt design, and CLIP-based filtering) to the domain of tableware contamination, enabling efficient and reliable synthetic data generation. (3)We provide extensive experimental validation, including ablation studies and deployment analyses, demonstrating both theoretical advances and practical feasibility.

\section{Related Work}
We review three technical domains central to DTGen: generative data augmentation, parameter-efficient fine-tuning, and vision-language models for data quality control.

\subsection{Generative Data Augmentation}
Data augmentation is a key strategy for improving generalization in deep models. Traditional geometric and color transformations are efficient but lack semantic diversity \cite{ICLR2024_3f1351a1}. Advances in GANs \cite{wang2017generative} and diffusion models \cite{10081412} have shifted attention toward generative augmentation, particularly for few-shot learning \cite{Lin_2023_CVPR}. Diffusion models are now state-of-the-art due to their superior fidelity and stability \cite{islam2025context}. Representative methods include DA-Fusion, which leverages pre-trained diffusion models for semantic variants \cite{ICLR2024_3f1351a1}, and DiffuseMix, which augments data via prompt-driven style mixing \cite{islam2024diffusemix}. However, these methods primarily perform global transformations akin to style transfer, limiting their ability to generate semantically novel content consistent with physical constraints.

DTGen advances beyond such approaches by combining LoRA-based domain specialization with structured prompt generation. This enables the creation of semantically novel contamination patterns that respect tableware-specific material and spatial attributes, moving from superficial style diversification to genuine content-level generation.

\subsection{Parameter-Efficient Fine-tuning}
Full fine-tuning of large diffusion models is computationally prohibitive \cite{lee2023multimodal}. Parameter-Efficient Fine-Tuning (PEFT) methods, especially LoRA \cite{hu2022lora}, address this by freezing most model weights while inserting small trainable low-rank matrices into key layers \cite{radford2021learning}. In text-to-image diffusion models such as Stable Diffusion \cite{rombach2022high}, LoRA typically targets cross-attention layers that align textual semantics with visual content \cite{Brownlee2024LoRA}. This approach enables efficient acquisition of new concepts with minimal data and resources, while LoRA modules remain lightweight \cite{lee2023multimodal}.

DTGen leverages LoRA for targeted adaptation to tableware contamination patterns. By constraining optimization on material textures, geometries, and contamination distributions, DTGen achieves deep domain alignment, laying the groundwork for high-quality contamination sample generation under few-shot conditions.

\subsection{Vision-Language Models for Data Filtering}
Ensuring the quality of synthetic data is critical, as low-quality or mislabeled samples can degrade downstream performance \cite{liang2022few}. Manual curation is costly and subjective, motivating automated solutions. CLIP \cite{radford2021learning}, trained via contrastive learning on large-scale image-text pairs, aligns visual and textual embeddings in a shared semantic space. This capability has been widely applied to dataset filtering and prompt-based retrieval \cite{hammoud2024synthclip}. SynthCLIP further demonstrated CLIP’s effectiveness even when trained on synthetic data, underscoring its utility in generative workflows.

In DTGen, CLIP serves as an automated quality control mechanism. By computing similarity scores between generated images and their prompts, DTGen programmatically filters out semantically inconsistent or artifact-prone outputs. This adaptive filtering ensures that the final synthetic dataset is both semantically aligned and reliable, a crucial prerequisite for robust few-shot learning \cite{nguyen2025provably}.

\section{Methodology}
This section introduces DTGen (Dirty Tableware Generation), a few-shot generative data augmentation framework tailored for tableware dirt recognition. DTGen expands a small set of annotated samples into large-scale, high-quality training data by leveraging domain-specific diffusion model fine-tuning, diverse dirt generation strategies, and robust tableware image filtering mechanisms.

\subsection{Fine-Grained Tableware Dirt Recognition Problem}
Unlike prior work that treats dirt recognition as a coarse binary classification task (clean vs. dirty), we redefine it as a multi-level fine-grained recognition problem. Given a tableware image $I \in \mathbb{R}^{H\times W\times 3}$, the goal is to predict its dirt state $y \in \mathcal{Y}$, where $\mathcal{Y}=\{c_0,c_1,...,c_K\}$ denotes $K+1$ categories ranging from completely clean to severely contaminated. Each category $c_k$ is characterized by three semantic attributes: dirt type $\mathcal{T}=\{grease, food\,residue, water\,stains,...\}$, contamination degree $\mathcal{S}=\{slight, moderate, severe\}$, and spatial distribution $\mathcal{D}=\{local, scattered, full\,coverage\}$. Formally, the fine-grained recognition task is defined as a mapping $f:\mathbb{R}^{H\times W\times 3}\rightarrow\mathcal{T}\times\mathcal{S}\times\mathcal{D}$. This formulation enables more precise assessment of contamination levels, providing valuable guidance for intelligent cleaning systems.

Under the few-shot setting, the central challenge is to learn a high-performance classifier $f_\theta:\mathbb{R}^{H\times W\times 3}\rightarrow\mathcal{Y}$ from a limited annotated dataset $\mathcal{D}_{real}=\{(I_i,y_i)\}_{i=1}^N$, where $N$ is the number of real samples and $N\ll|\mathcal{Y}|$, with $H$ and $W$ denoting image dimensions. The diversity and complexity of dirt patterns, combined with the scarcity of annotations, make conventional supervised approaches prone to severe overfitting, while simple augmentation methods such as geometric transformations fail to capture the semantic variability of dirt. To address this, we propose DTGen, which leverages generative models to transform scarce real samples into semantically rich, diverse, and quality-controlled synthetic data, thereby mitigating data scarcity in fine-grained tableware dirt recognition.

\begin{figure*}[t]
\centering
\includegraphics[width=0.85\textwidth]{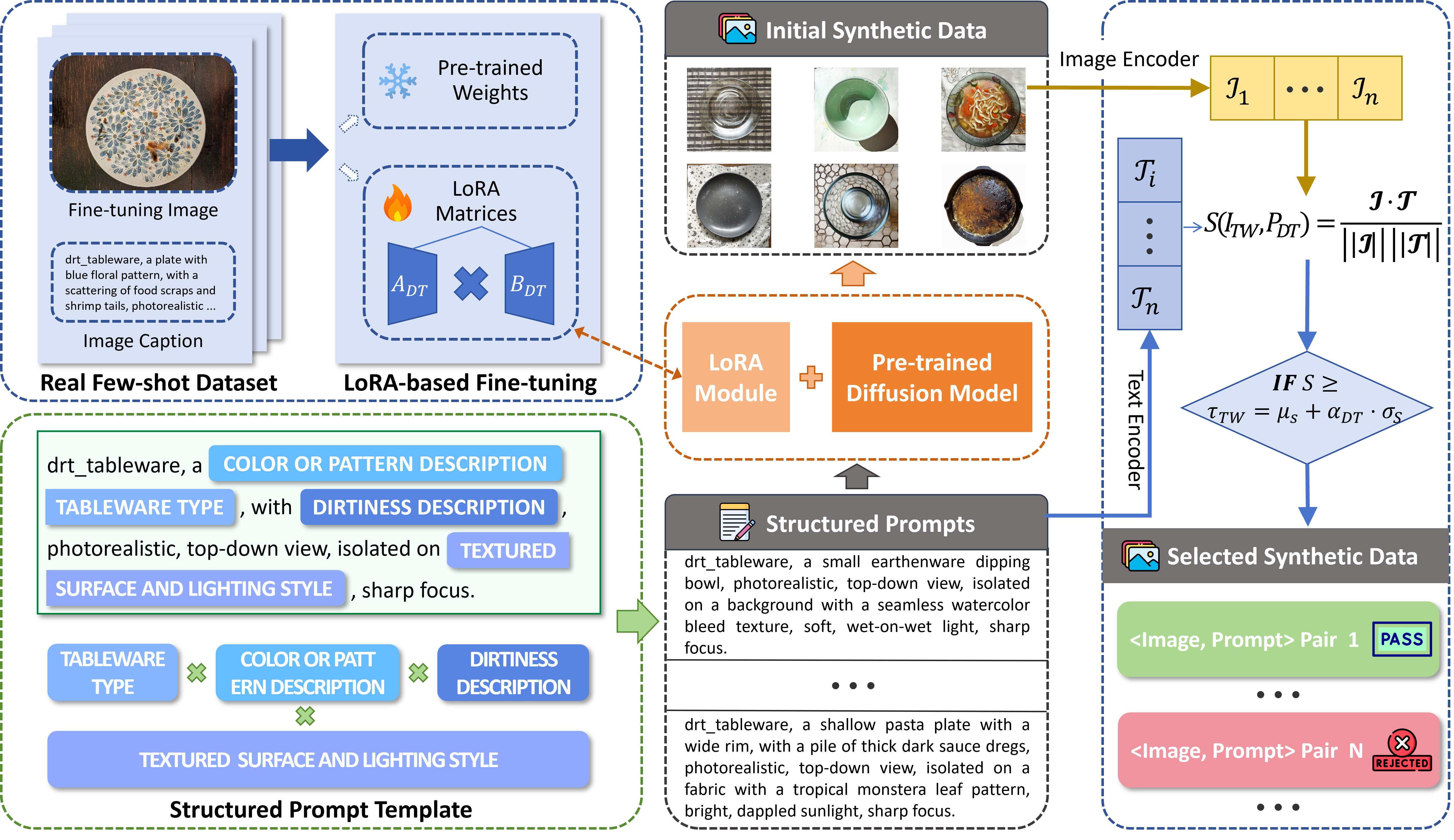} 
\caption{Flowchart of the proposed DTGen.}
\label{fig1}
\end{figure*}

\subsection{Proposed DTGen}
The central idea of DTGen is to establish an end-to-end pipeline that transforms scarce annotated tableware dirt data into high-performance recognition models. As illustrated in Figure \ref{fig1}, the framework progressively refines and enriches knowledge through four task-specific modules: (1) parameter-efficient domain adaptation that equips general generative models with tableware- and dirt-specific visual priors; (2) systematic expansion of dirt diversity via pattern-guided prompts; (3) image-text alignment evaluation to ensure both the realism of generated tableware and the accuracy of dirt semantics; and (4) fine-grained classifier training to acquire strong discriminative ability across diverse dirt patterns.

\subsubsection{Tableware Dirt Pattern Learning}
Although pre-trained diffusion models offer strong general generative abilities, they lack adequate representation of tableware-specific visual concepts and dirt patterns, making direct application ineffective. To address this, we adopt Low-Rank Adaptation (LoRA) \cite{hu2022lora} for parameter-efficient domain specialization. The key insight is that adapting deep networks to novel visual concepts primarily occurs in low-dimensional subspaces. Concretely, we fine-tune only the weight matrices $(W_Q,W_K,W_V,W_O)$ of the cross-attention layers in the U-Net backbone of Stable Diffusion \cite{rombach2022high}, thereby enabling the model to capture tableware textures, shapes, and diverse dirt characteristics. For a weight matrix $W_O \in \mathbb{R}^{d\times k}$, the update is formulated as:
\begin{equation}\label{eq1}
W=W_0+\Delta W_{DT}=W_0+B_{DT}\cdot A_{DT}
\end{equation}
where $d$ is the U-Net feature dimension, $r$ is the rank, and $B_{DT} \in \mathbb{R}^{d\times r}$, $A_{DT} \in \mathbb{R}^{r\times k}$ satisfy the low-rank constraint $r\ll \min(d,k)$. The optimization objective is designed to capture both tableware appearance and dirt distribution patterns, expressed as:
\begin{align}\label{eq2}
\mathcal{L}_{DTGen}=\mathbb{E}_{(I_{TW},t,\epsilon)}\left[\parallel\epsilon-
\epsilon_\theta(z_t,t,c_{DT})\parallel_2^2\right] \nonumber \\ 
+\lambda\parallel A_{DT}\parallel_F^2+\mu\parallel B_{DT}\parallel_F^2
\end{align}
where $z_t$ is the noisy latent representation of a tableware image, $c_{DT}$ is the textual description of tableware type and dirt state, and $\lambda,\mu$ are regularization coefficients to prevent overfitting. Through such targeted fine-tuning, the diffusion model not only retains its general generative capacity but also acquires a deeper understanding of tableware materials (e.g., ceramic gloss, glass transparency, metal reflection), dirt patterns (e.g., grease spread, food residue morphology, water stain edges), and typical lighting conditions in real usage scenarios.

\subsubsection{Diverse Dirt Data Generation}
To enable the fine-tuned diffusion model to produce large-scale synthetic datasets with diverse dirt patterns, we design a hierarchical semantic template specifically tailored for tableware dirt recognition. As illustrated in Figure \ref{fig1}, the template adopts a structured slot-filling mechanism, where each slot corresponds to a key visual factor influencing recognition: common tableware types (\texttt{TABLEWARE TYPE}), tableware styles or functions (\texttt{COLOR OR PATTERN DESCRIPTION}), dirt conditions (\texttt{DIRTINESS DESCRIPTION}), and imaging environments (\texttt{TEXTURED SURFACE AND LIGHTING STYLE}). Representative predefined options for each slot are provided in Table \ref{tab1}. 

By performing Cartesian product sampling over these semantic dimensions, we ensure uniform coverage of the dirt recognition space:
\begin{equation}\label{eq3}
\mathcal{P}_{DT\_diverse} = \prod_{i=1}^{K} \text{Uniform}(\mathcal{V}_{DT_i})
\end{equation}
where $\mathcal{V}_{DT_i}$ denotes the set of candidate values for the $i$-th slot. The template is easily extensible—updating slot options allows the generation of thousands of prompt sets while avoiding contradictions. Combined with the fine-tuned diffusion model, this approach can generate virtually unlimited photorealistic samples. 

This structured sampling strategy not only guarantees dataset scale, but more importantly ensures systematic control over multiple dimensions, including category balance, tableware diversity, material coverage, and contamination severity. As a result, it provides comprehensive coverage of realistic usage scenarios.

\begin{table}[t]
\centering
\renewcommand{\arraystretch}{1.2}
\begin{tabular}{m{2.8cm}<{\centering}m{4.2cm}<{\centering}}
\Xhline{1.5px}
    \textbf{Slot} & \textbf{Predefined Options}  \\
\Xhline{0.75px}
    \texttt{TABLEWARE TYPE} (6 items) & bowl; plate; teacup; ...  \\
\Xhline{0.75px}
    \texttt{COLOR OR PATTERN DESCRIPTION} (28 items) & a round white ceramic dinner; a heavy and durable melaminea; a dark and moody stoneware soup; ...  \\
\Xhline{0.75px}
    \texttt{DIRTINESS DESCRIPTION} (132 items) & a few crumbs scattered in the center; a significant pool of leftover tomato sauce; a whole slice of pizza fused to the surface, the cheese now a hard, moldy shell; ...  \\
\Xhline{0.75px}
    \texttt{TEXTURED SURFACE AND LIGHTING STYLE} (16 items) & a ceramic tile with an intricate Mandala pattern, soft; a wallpaper with a William Morris print pattern, soft; bright kitchen light; even light; classical light; ... \\
\Xhline{1.5px}
\end{tabular}
\caption{Structured prompt template for diverse dirty tableware image generation.}
\label{tab1}
\end{table}

\subsubsection{High-Quality Sample Filtering}
Generative models often display large quality variations when synthesizing tableware dirt images, especially for complex dirt patterns. Failures typically manifest as unrealistic tableware shapes, incorrect dirt distributions, or implausible material appearances. Such low-quality samples introduce noisy labels during training, substantially degrading classifier performance. To address this, we leverage the cross-modal alignment capability of the CLIP model \cite{radford2021learning} to design an automated quality assessment and filtering mechanism tailored for tableware images. 

Given a generated image–prompt pair $(I_{TW},P_{DT})$, we extract visual and textual embeddings using CLIP’s encoders and compute their cosine similarity:
\begin{equation}\label{eq4}
\fontsize{8.5}{12}\selectfont
S(I_{TW},P_{DT})=\frac{\text{CLIP}_{image}(I_{TW})\cdot\text{CLIP}_{text}(P_{DT})}
{\parallel\text{CLIP}_{image}(I_{TW})\parallel\parallel\text{CLIP}_{text}(P_{DT})
\parallel}
\end{equation}

Since different tableware types and dirt severities exhibit varying generation difficulties (e.g., slight water stains on transparent glass are harder to synthesize than heavy oil stains on stainless steel), we introduce an adaptive thresholding strategy:
\begin{equation}\label{eq5}
\tau_{TW}=\mu_S+\alpha_{DT}\cdot\sigma_S
\end{equation}
where $\mu_S$ and $\sigma_S$ denote the mean and standard deviation of similarity scores across samples generated by the current prompt, and $\alpha_{DT}$ is a control coefficient adjusted according to dirt complexity. Empirically, we set $\alpha_{DT}\approx 1.5$ based on experimental validation and human evaluation. 

This adaptive mechanism dynamically calibrates filtering criteria, ensuring effective noise suppression while retaining valid dirt samples that might otherwise be discarded. From an information-theoretic perspective, the filtering process maximizes mutual information between generated data and real-world dirt distributions, thereby preserving high fidelity and a favorable signal-to-noise ratio for downstream classifier training.

\subsubsection{Tableware Dirt Classifier Enhancement}
With the high-quality synthetic dataset $D_{syn}$ obtained from the previous steps, we construct and train the final tableware dirt recognition classifier. Any standard image classification backbone can serve as the base model; by replacing its output layer with a multi-classification head, it can be adapted to our fine-grained dirt recognition task. 

Classifier training follows standard protocols, including the AdamW optimizer, a step-based learning rate scheduler, and appropriate batch sizes. We train exclusively on $D_{syn}$ without relying on real images. After convergence, the enhanced classifier demonstrates stronger discriminative ability across diverse dirt patterns, yielding not only higher recognition accuracy but also improved robustness to variations in tableware materials, illumination conditions, and dirt complexities. This enhanced generalization capability makes the model more reliable in real-world intelligent cleaning scenarios.

\subsection{Time and Space Complexity Analysis}
In terms of computational complexity, since the number of real samples $N$ is far smaller than the number of generated samples $M_{TW}$ and the rank $r$ is much smaller than the diffusion backbone dimension $d$, the overall time cost of DTGen across its four stages (pattern learning, data generation, filtering, and classifier training) can be approximated as $\mathcal{O}(M_{TW}\cdot K\cdot H\cdot W\cdot d^2)$. This complexity is dominated by the data generation stage and scales linearly with the number of generated samples, demonstrating favorable scalability. 

For space complexity, DTGen benefits from LoRA’s low-rank decomposition, which requires only $\mathcal{O}(r\cdot d)$ additional parameters—negligible compared to the original model parameters $\Theta_{base}$. Together with gradient checkpointing and streaming strategies, the overall space complexity can be expressed as $\mathcal{O}(r\cdot d+M_{DT}\cdot H\cdot W)$. 

In summary, DTGen achieves strong computational and storage efficiency while maintaining scalability and high-quality data generation, providing both theoretical support and practical feasibility for large-scale deployment in intelligent tableware cleaning systems.

\section{Experiments and Results}
To rigorously evaluate the effectiveness of DTGen, we conducted a series of comparative experiments designed to assess its capability in addressing extreme data scarcity for fine-grained tableware dirt recognition.

\subsection{Experimental Setup}
\begin{itemize}
  \item \textbf{Dataset:} We adopted the Cleaned vs Dirty V2 dataset \cite{platesv2}, the only publicly available benchmark for tableware dirt classification, containing $784$ high-resolution images of ceramic plates and metal bowls under diverse lighting conditions. To simulate extreme few-shot scenarios, $40$ images ($20$ clean, $20$ dirty) were used for training, while the remaining $744$ images served as an independent test set.
  
  \item \textbf{Implementation:} DTGen was built on Stable Diffusion 3.5 Large. LoRA modules with rank $r=8$ were trained for $1000$ steps on the $40$ training samples to capture contamination patterns without overfitting. Structured prompts generated $3600$ synthetic images, which were filtered by CLIP (openai/clip-vit-base-patch32) using cosine similarity thresholds, yielding $3297$ high-quality samples as the final training set $D_{selected}$. For classification, ImageNet-pretrained ResNet-50 models with task-specific output layers were trained solely on $D_{selected}$.
  
  \item \textbf{Metrics:} Performance was evaluated using accuracy, precision, recall, and F1-score, providing a balanced assessment of classification correctness, positive prediction reliability, and coverage.
\end{itemize}

\subsection{Baselines}
As no existing methods target fine-grained tableware dirt recognition, we construct two baselines based on general vision learning techniques and assess the necessity of DTGen components through two ablation baselines:

\begin{itemize}
  \item \textbf{Few-Shot:} Training a ResNet-50 classifier directly on $40$ real few-shot samples \cite{he2016deep}, serving as the lower-bound performance reference under severe data scarcity.
  \item \textbf{Traditional Data Augmentation (TDA):} Applying conventional geometric transformations \cite{wang2017effectiveness} ($4$ geometric operations, $6$ rotations, $5$ color jitters) on the Few-Shot dataset, producing $4800$ augmented images to assess the effectiveness of traditional augmentation.
  \item \textbf{DTGen (w/o LoRA):} Generating $3600$ synthetic images with high-quality filtering using a general diffusion model without domain specialization, isolating the contribution of tableware dirt pattern learning.
  \item \textbf{DTGen (w/o CLIP):} Generating $3600$ synthetic images with LoRA fine-tuned models but without high-quality filtering, highlighting the impact of sample quality control.
\end{itemize}

\subsection{DTGen Performance Evaluation}
Table \ref{tab2} compares the binary classification performance of DTGen with several baseline methods on the held-out test set. The results highlight the clear limitations of traditional approaches. The Few-Shot method performed the worst, achieving only $0.65$ accuracy. Although its recall reached $1.00$, the very low precision indicates that the model degenerated into a trivial solution, predicting nearly all samples as ``dirty''—a typical symptom of severe overfitting. The TDA method achieved moderate improvements, with accuracy of $0.81$ and recall of $0.99$, yet still suffered from similar issues, showing that conventional augmentation cannot fundamentally address the lack of dirt pattern diversity.

\begin{table}[t]
\centering
\small
\renewcommand{\arraystretch}{2}
\begin{tabular}{ccccc}
\Xhline{1.5px}
    \textbf{\makecell{Training\\Scheme}} & \textbf{Precision} & \textbf{Recall} & \textbf{F1-Score}& \textbf{Accuracy} \\
\Xhline{0.75px}
   Few-Shot & $0.65$ & $\mathbf{1.00}$ & $0.79$ & $0.65$ \\

   TDA & $0.81$ & $0.99$ & $0.89$ & $0.84$ \\

   \makecell{DTGen\\(w/o LoRA)} & $0.68$ & $0.99$ & $0.81$ & $0.69$  \\

   \makecell{DTGen\\(w/o CLIP)} & $0.89$ & $0.91$ & $0.92$ & $0.92$ \\

   DTGen & $\mathbf{0.90}$ & $0.92$ & $\mathbf{0.93}$ & $\mathbf{0.93}$ \\
\Xhline{1.5px}
\end{tabular}
\caption{Performance comparison between DTGen and baseline methods on tableware dirt binary classification tasks.}
\label{tab2}
\end{table}

\begin{table}[t]
\centering
\small
\renewcommand{\arraystretch}{2}
\begin{tabular}{ccccc}
\Xhline{1.5px}
    \textbf{\makecell{Training\\Scheme}} & \textbf{Precision} & \textbf{Recall} & \textbf{F1-Score}& \textbf{Accuracy} \\
\Xhline{0.75px}

   TDA & $0.69$ & $0.72$ & $0.70$ & $0.71$ \\

   DTGen & $\mathbf{0.86}$ & $\mathbf{0.87}$ & $\mathbf{0.86}$ & $\mathbf{0.86}$ \\
\Xhline{1.5px}
\end{tabular}
\caption{Performance comparison between DTGen and TDA on tableware dirt three-class classification tasks.}
\label{tab3}
\end{table}

Ablation studies further confirm the contribution of each DTGen component. DTGen (w/o LoRA) yielded only marginal gains, with an F1-score of $0.81$, suggesting that generic synthetic data alone is insufficient for performance improvement. In contrast, DTGen (w/o CLIP) achieved a substantial increase, reaching an F1-score of $0.92$, underscoring the pivotal role of LoRA-based domain adaptation in generating visually and semantically aligned data. The complete DTGen achieved the best result with an F1-score of $0.93$, highlighting CLIP’s effectiveness as a ``semantic guardrail'' for filtering noisy samples and ensuring training set quality. Moreover, since DTGen operates independently of traditional augmentation, it can be seamlessly combined with geometric transformations, offering an additional pathway for further performance enhancement.

\subsection{Potential of Fine-Grained Recognition}
To further assess DTGen’s applicability in refined dirt recognition tasks, we manually re-annotated the held-out test set into three categories (clean, lightly dirty, heavily dirty). This setup allows us to validate both the semantic richness of DTGen-generated data and the model’s fine-grained discriminative ability. As shown in Table \ref{tab3}, DTGen achieved an accuracy of $0.86$, substantially surpassing TDA’s $0.71$. More importantly, DTGen maintained balanced performance across all categories, avoiding the severe misclassification issues observed in traditional methods. These results demonstrate DTGen’s strong potential for extension to more fine-grained recognition tasks. By providing diverse synthetic samples, DTGen enables models to capture subtle visual distinctions among different contamination levels, establishing a solid foundation for high-precision systems capable of distinguishing multiple attributes such as stain types, contamination degrees, and spatial distributions.

\subsection{Generated Data Visualization Analysis}
\begin{figure}[t]
\centering
\subfigure[Samples retained through high-quality filtering]{
\includegraphics[width=0.9\linewidth]{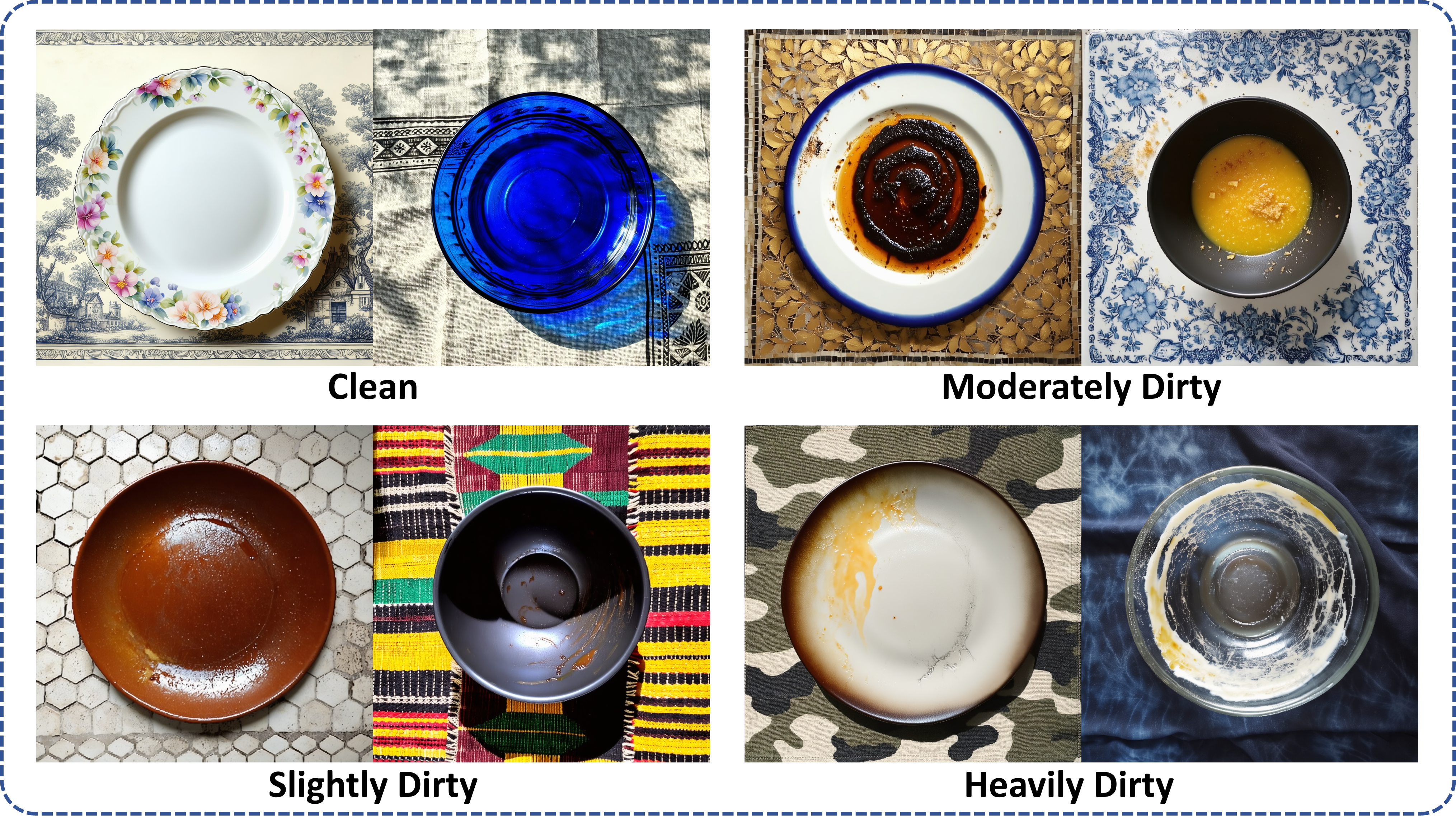}
\label{fig2a}
}
\subfigure[Low-quality samples rejected by quality filtering]{
\includegraphics[width=0.9\linewidth]{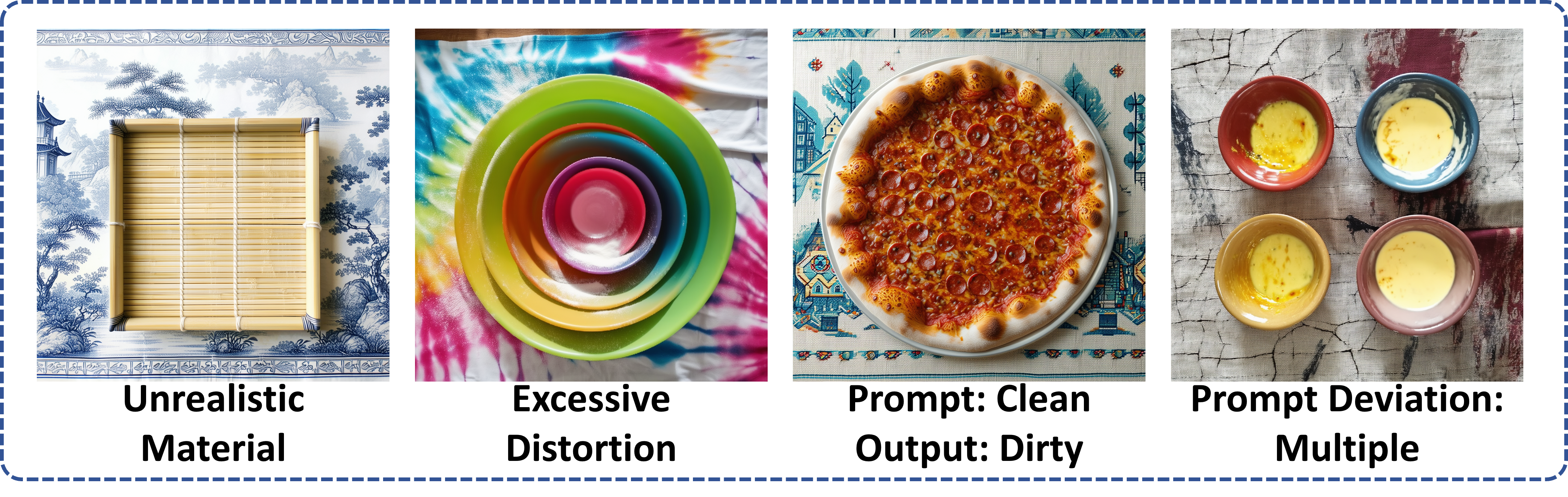}
\label{fig2b}
}
\caption{Visualization of DTGen generated data.}
\label{fig2}
\end{figure}

To provide an intuitive understanding of DTGen’s effectiveness, we conducted a qualitative analysis of the generated data. Figure \ref{fig2} compares samples retained after high-quality filtering with those discarded. The retained samples (Figure \ref{fig2a}) exhibit high diversity and realism across tableware types, materials, lighting conditions, and stain morphology, capturing subtle features such as ``dried sauce'' and ``greasy streaks'' that supply rich cues for classifier learning. In contrast, the discarded samples (Figure \ref{fig2b}) illustrate common generative failures, including distorted object shapes, poor adherence to prompts, and unnatural texture artifacts. This side-by-side comparison highlights the necessity and effectiveness of the filtering mechanism as a critical quality assurance step.

\section{Path to Deployment}
Industrial deployment of DTGen requires addressing key technical challenges, including edge resource constraints, model optimization, and system integration. Based on market demand analysis for intelligent dishwashers, we outline a two-stage deployment pathway.

\begin{itemize}
  \item \textbf{Edge device model optimization:} Standard ResNet-50 models (200+MB) are unsuitable for typical dishwasher controllers, which usually employ low-end processors with less than 1GB memory. Thus, models must be compact and support fast inference. We adopt a three-step optimization strategy: (1) knowledge distillation to transfer knowledge from ResNet-50 teacher models to lightweight MobileNetV3 \cite{howard2019searching} students, achieving substantial compression while retaining accuracy; (2) INT8 quantization \cite{intel2023x86quant} to further reduce size; and (3) structured pruning \cite{dong2019network} to remove redundancy, targeting real-time recognition on resource-limited hardware.
  
  \item \textbf{Intelligent system integration:} The optimized DTGen model communicates with dishwasher controllers via CAN bus, enabling real-time intelligent cleaning decisions. Cleaning modes are automatically adapted to recognition results: energy-saving cycles for light dirt, standard programs for moderate dirt, and intensive cleaning with extra rinses for heavy dirt. Detergent usage is also optimized, e.g., increasing degreaser for grease and adjusting enzyme formulations for protein residues. This integration strategy improves cleaning effectiveness and energy efficiency while enhancing user experience.
\end{itemize}

\section{Conclusion}
We proposed DTGen, a generative data augmentation framework designed to address data scarcity in tableware contamination recognition. By transforming only $40$ few-shot samples into over $3000$ high-quality synthetic images, DTGen significantly enhances fine-grained recognition performance. Its key novelty lies in integrating LoRA-based domain adaptation, contamination-aware prompt design, and CLIP-driven quality filtering, together forming an effective solution for tableware hygiene monitoring and a general paradigm for few-shot industrial vision tasks. Our results demonstrate the potential of generative AI to advance traditional industrial applications toward intelligent transformation, laying the foundation for next-generation automated hygiene monitoring systems.

Nevertheless, DTGen still faces challenges, including high computational cost and limited coverage of extreme contamination. Future work will explore more efficient generative architectures to reduce resource demands, extend to multi-class recognition for richer decision support, and integrate sensor feedback for fully autonomous cleaning. With the rapid progress of generative AI, DTGen also shows strong potential for broader applications such as food safety inspection and medical device cleaning validation.

\section*{Acknowledgments}
This work was supported in part by the National Natural Science Foundation of China under Grants 62402249, 62262046, and 42161070, in part by the Science \& Technology Plan Project of Inner Mongolia A. R. of China under Grants 2022YFSJ0027, and 2023KJHZ0016, in part by the Ordos Science \& Technology Plan Grant YF20240029, in part by the fund of supporting the reform and development of local universities (Disciplinary construction), and in part by the fund of First-class Discipline Special Research Project of Inner Mongolia A. R. of China under Grant YLXKZX-ND-036.

\newpage

\bibliography{aaai2026}

\end{document}